\documentclass[12pt, a4paper]{article}
\usepackage[utf8]{inputenc}
\usepackage[english]{babel}
\usepackage{graphicx}
\usepackage{amsmath}
\usepackage{amssymb}
\usepackage{cite}
\usepackage{hyperref}
\usepackage{booktabs}
\usepackage{tabularx, booktabs}

\usepackage{multirow}
\usepackage{algorithm}
\usepackage{algorithmic}
\usepackage{geometry}
\usepackage{setspace}
\title{\textbf{Combating data scarcity in recommendation services: Integrating cognitive types of VARK and neural network technologies (LLM)}}

\author{
Nikita V. Zmanovskii\\
\\
Russian Biotechnological University (ROSBIOTECH), Moscow, Russia\\
\texttt{zmanovskiy.n.v@gmail.com}
}

\date{}

\begin{document}

\maketitle

\begin{abstract}
Cold start scenarios present fundamental obstacles to effective recommendation generation, particularly when dealing with users lacking interaction history or items with sparse metadata. This research proposes an innovative hybrid framework that leverages Large Language Models (LLMs) for content semantic analysis and knowledge graph development, integrated with cognitive profiling based on VARK (Visual, Auditory, Reading/Writing, Kinesthetic) learning preferences. The proposed system tackles multiple cold start dimensions: enriching inadequate item descriptions through LLM processing, generating user profiles from minimal data, and dynamically adjusting presentation formats based on cognitive assessment. The framework comprises six integrated components: semantic metadata enhancement, dynamic graph construction, VARK-based profiling, mental state estimation, graph-enhanced retrieval with LLM-powered ranking, and adaptive interface design with iterative learning. Experimental validation on MovieLens-1M dataset demonstrates the system's capacity for personalized recommendation generation despite limited initial information. This work establishes groundwork for cognitively-aware recommendation systems capable of overcoming cold start limitations through semantic comprehension and psychological modeling, offering personalized, explainable recommendations from initial user contact.
\end{abstract}


\section{Introduction}

The exponential growth of digital content across platforms necessitates sophisticated recommendation mechanisms to assist users in navigating information overload. Modern recommender systems have become integral to user experience in domains ranging from entertainment streaming to online education and e-commerce. However, despite substantial research advances, these systems continue facing critical challenges that significantly impact their real-world effectiveness.

\subsection{The Challenge of Cold Start}

Among persistent difficulties in recommendation research, the cold start problem stands as particularly intractable. This challenge manifests when systems encounter users without sufficient interaction history, newly introduced items lacking ratings, or entirely new platforms without accumulated data. Traditional collaborative filtering approaches, which form the backbone of many successful systems, fundamentally rely on historical interaction patterns. When such patterns are unavailable, these methods fail to generate meaningful suggestions, potentially causing negative first impressions that may permanently affect user engagement.

The impact extends beyond technical performance metrics. Research indicates that users experiencing poor initial recommendations demonstrate significantly higher abandonment rates and reduced long-term platform engagement. This creates a critical window during which systems must perform effectively despite minimal available information about user preferences or item characteristics.

\subsection{Limitations in Current Solutions}

Existing approaches to cold start scenarios typically employ one of several strategies, each carrying inherent weaknesses. Content-based filtering attempts matching items through feature similarity but requires comprehensive, structured metadata often unavailable in practice. These systems also struggle capturing complex semantic relationships beyond surface-level attributes and tend toward recommendation homogeneity, limiting diversity and serendipitous discovery.

Demographic-based methods leverage user population segments to generate recommendations, but rely on broad generalizations rather than individual preference nuances. Knowledge-based systems utilize domain expertise and explicit requirements but demand substantial manual effort for knowledge encoding and often lack scalability across different domains.

Recent deep learning approaches, including meta-learning and transfer learning methods, show promise but typically require extensive training data from similar domains and may not generalize effectively to truly novel scenarios. Generative models can synthesize user preferences or item representations but often struggle producing realistic, diverse recommendations capturing real preference complexity.

\subsection{Unaddressed Dimensions}

Beyond basic cold start challenges, current systems fail addressing several critical personalization aspects. Individual cognitive differences in information processing remain largely ignored—users vary significantly in their preferred information reception modes, with some responding better to visual presentations while others prefer textual or interactive formats. Existing systems typically employ uniform presentation styles regardless of these fundamental cognitive variations.

Temporal fluctuations in user cognitive states represent another overlooked dimension. User capacity and willingness to engage with different content types varies based on factors including time of day, fatigue levels, available time, and current objectives. Most systems treat users as having static preferences and cognitive capabilities, failing to adapt recommendations to these dynamic contextual factors.

Current systems also underutilize the rich semantic content inherent in items. Most rely on sparse, structured metadata that inadequately captures the complex semantic relationships and nuanced characteristics of content, particularly for complex domains like educational materials where understanding prerequisites, difficulty progression, and learning objectives proves crucial.

\subsection{Research Contributions}

This paper introduces a comprehensive hybrid architecture addressing these multifaceted challenges through integration of advanced technologies and psychological principles:

\textbf{Semantic Enhancement via LLMs:} We employ Large Language Models to transform incomplete, unstructured metadata into semantically rich item profiles, extracting entities, relationships, difficulty assessments, prerequisites, and target audiences.

\textbf{Dynamic Graph Infrastructure:} Based on enriched profiles, we construct multi-relational knowledge graphs capturing complex semantic relationships between items and users, enabling sophisticated graph neural network-based algorithms functioning effectively with minimal interaction history.

\textbf{VARK-Based Cognitive Profiling:} Integration of the VARK learning styles framework enables psychologically grounded user profiles capturing individual cognitive preferences for personalized recommendation presentation.

\textbf{Mental State Modeling:} A cognitive simulator estimates users' current mental states from contextual signals (time, session duration, device type, stated goals), enabling dynamic content and presentation adaptation.

\textbf{Zero-Shot Capabilities:} By combining LLM-generated preference hypotheses with VARK profiles, the system generates personalized recommendations from minimal input data.

\textbf{Adaptive Presentation with Learning:} Recommendations include LLM-generated personalized explanations with dynamically adapted visual presentations, while the system continuously refines knowledge graphs and user profiles through interaction feedback.

\subsection{Paper Structure}

Section 2 reviews relevant literature across cold start solutions, LLM-based systems, knowledge graph methods, and cognitive modeling. Section 3 details the hybrid architecture and its six core modules. Section 4 presents experimental methodology including datasets, baselines, and metrics. Section 5 reports and analyzes results. Section 6 discusses implications and future directions. Section 7 concludes with key findings summary.

\section{Related Work}

This section surveys relevant research across four domains: traditional cold start approaches, LLM-based recommendation systems, knowledge graph methodologies, and cognitive modeling in personalization.

\subsection{Traditional Approaches to Cold Start}

Content-based methods represent early cold start solutions, analyzing item features to recommend similar items to user preferences. While effective for new item scenarios, these approaches demonstrate limited diversity and require rich metadata often unavailable practically. Demographic approaches utilize user population characteristics for initial recommendations but make broad generalizations failing to capture individual nuances. Knowledge-based systems employ domain expertise but require significant manual effort and struggle with cross-domain scalability.

\subsection{Deep Learning Innovations}

Recent deep learning advances have introduced novel cold start approaches. Meta-learning methods train models for rapid adaptation to new users or items with minimal data, though typically requiring substantial training data from similar contexts. Transfer learning leverages knowledge from data-rich domains but faces challenges identifying appropriate source domains and effective knowledge transfer. Variational autoencoders and generative models show promise generating synthetic preferences but often struggle with realistic, diverse output.

\subsection{LLM-Based Systems}

Large Language Models have opened new cold start possibilities. Wu et al. \cite{lmprior} demonstrate that language models can incorporate prior item knowledge to overcome cold start challenges through semantic item representations enabling recommendations for new items without interaction history. Zhang et al. \cite{acdr} propose adaptive candidate retrieval with dynamic knowledge graph construction, using LLMs for entity and relationship extraction from item descriptions. Wang et al. \cite{coldstart_bias} investigate potential biases in LLM-based systems during cold start, revealing unexpected biases in item descriptions and suggesting careful prompt engineering and mitigation strategies.

\subsection{Knowledge Graph Methods}

Knowledge graphs have emerged as powerful tools addressing data sparsity and cold start problems. These structured representations capture entities and relationships, enabling sophisticated preference reasoning. Graph embedding methods learn dense vector representations preserving graph structure for predicting missing links. Graph neural networks demonstrate impressive results through iterative message passing, enabling effective reasoning about indirect relationships particularly suitable for sparse connection scenarios. Multi-relational graphs incorporating various relationship types provide richer recommendation context but typically require significant manual effort or extensive structured data.

\subsection{Cognitive Modeling and VARK}

The VARK model categorizes learners by preferred information processing modes: Visual, Auditory, Reading/Writing, and Kinesthetic. Originally developed for educational contexts, VARK has been applied across domains requiring personalized information delivery. Klašnja-Milićević et al. \cite{vark} demonstrate VARK-based recommendation effectiveness for adaptive e-learning, showing that aligning content presentation with learner preferences significantly improves engagement and outcomes. Aziz et al. \cite{hybrid} propose hybrid attribute-based systems for personalized e-learning addressing cold start through combined content-based and collaborative approaches incorporating learning style preferences.

\subsection{Research Gaps}

Despite progress, several gaps persist: limited integration of psychological principles with modern AI for comprehensive cold start solutions, insufficient exploration of temporal cognitive state variations in recommendation adaptation, underexplored combination of LLM semantic understanding with knowledge graph reasoning, limited work on adaptive presentation matching user cognitive preferences, and few end-to-end solutions addressing both user-side and item-side cold start with cognitive adaptation. This work addresses these gaps through a comprehensive hybrid architecture synergistically combining LLMs, knowledge graphs, and cognitive modeling.

\section{Methodology}

\subsection{Problem Formulation}

We define the cold start recommendation problem formally:

\textbf{Input:} Item set $\mathcal{I} = \{i_1, ..., i_n\}$ with sparse metadata; new user $u$ with minimal demographics and no interaction history; contextual information $C = \{c_1, ..., c_k\}$ about current session.

\textbf{Output:} Ranked item list $R = [i_{r_1}, ..., i_{r_K}]$ personalized for $u$; presentation format $F$ adapted to cognitive profile; natural language explanations $E = \{e_1, ..., e_K\}$ for recommendations.

\textbf{Constraints:} Zero historical interaction data; minimal, potentially unstructured metadata; real-time recommendation generation; adaptability to cognitive state and preferences.

Our objective maximizes recommendation quality (relevance, diversity, serendipity) while providing adaptive user experiences accounting for individual cognitive characteristics and current mental states.

\subsection{Architecture Overview}

The hybrid architecture consists of six interconnected modules in a continuous cycle:

\begin{enumerate}
\item \textbf{Metadata Enrichment}: Transforms raw metadata into semantically rich profiles via LLMs
\item \textbf{Knowledge Graph Construction}: Builds and maintains dynamic multi-relational graphs
\item \textbf{User Profiling}: Creates comprehensive profiles incorporating VARK learning styles
\item \textbf{Cognitive State Modeling}: Assesses current mental state and cognitive capacity
\item \textbf{Retrieval and Ranking}: Generates candidates and ranks using graph methods and LLM reasoning
\item \textbf{Presentation and Learning}: Adapts recommendation presentation and continuously learns from interactions
\end{enumerate}

\subsection{Module 1: LLM-Based Metadata Enrichment}

This module addresses incomplete, unstructured item metadata through LLM-based semantic enhancement.

For each item $i \in \mathcal{I}$, we construct semantic profile $P_i$ including: key entities (named entities, concepts, themes); relationships between entities; complexity level (estimated difficulty); required context (prerequisites); target audience characteristics; VARK alignment (alignment with different learning/consumption styles).

We employ carefully designed prompts guiding LLM information extraction:

\begin{verbatim}
Analyze this item and provide comprehensive semantic profile:
Title: [item_title]
Description: [item_description]
Metadata: [structured_metadata]

Provide:
1. Key entities and concepts with descriptions
2. Entity relationships
3. Complexity level (1-5) with justification
4. Required background knowledge
5. Target audience characteristics
6. Learning style alignment (V/A/R/K)
\end{verbatim}

The LLM output is parsed extracting structured entities and relationships. Each entity and relationship receives natural language description for interpretability and dense vector embedding for semantic similarity computation using pre-trained sentence transformers.

For each item, we assess VARK alignment: Visual (diagrams, charts, infographics present?), Auditory (audio content, discussions?), Reading/Writing (text-based format?), Kinesthetic (hands-on activities, interactive elements?). This enables matching items to users based on cognitive preferences beyond topic relevance.

\subsection{Module 2: Dynamic Knowledge Graph Construction}

This module transforms enriched semantic profiles into structured, queryable knowledge graphs.

Our graph $G = (V, E, T_V, T_E)$ consists of: nodes $V = V_I \cup V_E \cup V_U$ (item nodes, entity nodes, user nodes); edges $E$ (typed relationships); node types $T_V$; edge types $T_E$.

Each node $v \in V$ has attributes: unique identifier, type, name, natural language description, dense vector embedding $e_v \in \mathbb{R}^d$, and type-specific structured attributes. Each edge $(v_i, v_j, t) \in E$ has: relationship type, description, embedding, and weight/confidence score.

Graph construction follows: create item nodes with semantic profile attributes; extract entities creating or linking to existing entity nodes; establish edges between items and entities based on relationships; identify cross-item connections through shared entities and semantic similarity; perform multi-hop reasoning to infer implicit relationships.

We implement the knowledge graph using: graph database (Neo4j) for structure storage and efficient queries; vector database (FAISS) for embedding indexing and fast similarity search; text index (Elasticsearch) for full-text search over descriptions.

The graph dynamically evolves: new items are processed through enrichment and integrated; user interactions create new edges and strengthen existing ones; entity importance updates based on engagement; edge weights adjust based on predictive performance.

\subsection{Module 3: VARK-Based User Profiling}

This module creates comprehensive user profiles capturing explicit preferences and cognitive characteristics.

For new user $u$, we collect basic information (name, age, location, education) and primary usage goal $g \in \{$purchase, entertainment, research, learning$\}$.

Users complete a 16-question VARK questionnaire identifying dominant learning style. Based on responses, we compute VARK score vector:
$$\text{VARK}(u) = [v, a, r, k] \in [0,1]^4$$
where components represent preference strengths with $\sum_{i} \text{VARK}(u)_i = 1$.

The user profile encodes as:
$$P_u = \{d_u, g_u, \text{VARK}_u, e_u\}$$
where $d_u$ represents demographic features, $g_u$ is usage goal (one-hot encoded), $\text{VARK}_u$ is the score vector, and $e_u \in \mathbb{R}^d$ is learned user embedding (initialized randomly, refined through interactions).

A user node $v_u$ is created in the knowledge graph with edges to demographic/goal entities, VARK preference entities, and placeholder edges to items (filled through interactions).

\subsection{Module 4: Cognitive State Modeling}

This module models users' current mental states and cognitive capacity from contextual signals.

For each session, we collect: temporal context (time of day, day of week); device context (device type, platform); session context (duration, items viewed, interaction pace); explicit signals (stated goal, available time if provided).

Based on these signals and VARK profile, we estimate:

\textbf{Cognitive Load Capacity:} Expected mental capacity from time and session duration using:
$$\text{capacity}(t, s) = \alpha_t \cdot \beta_s$$
where $\alpha_t$ represents time-of-day factor and $\beta_s$ represents session fatigue factor.

\textbf{Attention Span:} Estimated attention from device and browsing pattern (mobile + fast browsing suggests short attention; desktop + careful examination suggests longer attention).

\textbf{Preferred Complexity:} Based on stated goal and current capacity (learning + high capacity can handle complex content; entertainment + low capacity prefers accessible content).

\textbf{Optimal Presentation Mode:} Combination of VARK preference and current context (visual learner + mobile emphasizes images; kinesthetic learner + high attention suggests interactive content).

We represent cognitive state as vector:
$$C_u = [cap, att, comp, pres] \in \mathbb{R}^4$$
where components represent cognitive capacity (0-1), attention span estimate (0-1), preferred complexity level (0-1), and presentation mode preferences (multi-hot VARK encoding).

\subsection{Module 5: Graph-Based Retrieval and LLM Ranking}

This module generates candidate items through graph-based retrieval and ranks via LLM reasoning.

We employ multi-strategy candidate generation:

\textbf{Semantic Similarity:} Query vector database for items with embeddings similar to user profile:
$$\mathcal{C}_1 = \text{TopK}_{i \in \mathcal{I}} \text{sim}(e_u, e_i)$$

\textbf{Entity-Based Retrieval:} If profile includes specific interests or goal suggests certain entities, retrieve items connected to those entities in graph.

\textbf{VARK-Aligned Retrieval:} Filter and prioritize items matching user VARK preferences.

\textbf{Cognitive State Filtering:} Filter candidates based on cognitive capacity and complexity alignment.

Final candidate pool combines these strategies, typically sized at 500-1000 items.

Given candidate pool $\mathcal{C}$ and profile $P_u$, we use LLM for ranking. The prompt includes user profile summary, candidate items with semantic profiles, and ranking criteria: relevance to goal, VARK alignment, appropriate complexity, diversity, and serendipity potential. The LLM generates ranked list $R = [i_1, ..., i_{20}]$ with justifications.

For robustness, we employ hybrid ranking: primary LLM-based ranking as described; fallback learned ranking model combining TF-IDF similarity, graph-based features, VARK alignment scores, and collaborative filtering signals if available.

\subsection{Module 6: Adaptive Presentation and Learning}

This module handles recommendation presentation and continuous improvement through feedback.

For each recommended item $i \in R$, we generate personalized explanation using LLM, producing 2-3 sentence explanations referencing specific profile aspects, highlighting VARK alignment, mentioning relevant entities or prerequisites, and maintaining natural, engaging tone.

The presentation format dynamically adapts to VARK profile and cognitive state: visual learners receive emphasized thumbnails, color coding, and visual indicators; auditory learners see highlighted audio previews and discussion features; reading/writing learners get detailed text descriptions and excerpts; kinesthetic learners see emphasized hands-on components and interactive demos. When cognitive capacity is low, we reduce information density, simplify language, limit initial options, and emphasize lighter content.

The system tracks all interactions: implicit feedback (click-through rate, viewing time, scroll depth, skip actions) and explicit feedback (ratings, comments, wishlist actions, completions).

User interactions drive continuous improvement: user embedding refinement updates $e_u$ based on interacted items; knowledge graph updates create edges between user and items, increase weights for positive interactions, and strengthen entity connections; VARK profile refinement gradually adjusts if user consistently engages with mismatched content; cognitive model calibration adjusts capacity estimation and complexity preferences based on engagement patterns; recommendation model updates through periodic retraining using accumulated interaction data.

To avoid filter bubbles, we employ the SerenEva framework \cite{sereneva}: tracking recommendations both relevant and novel, periodically injecting diverse potentially serendipitous items, evaluating recommendation sets for balance, and using LLM to generate and test serendipity hypotheses.

\section{Experimental Setup}

\subsection{Dataset}

We evaluate on MovieLens-1M dataset containing 1,000,209 ratings from 6,040 users on 3,706 movies with ratings on 1-5 scale, user demographics (age, gender, occupation), and movie metadata (title, genres, release year).

For cold start simulation: randomly partition users into training (80\%) for initial graph building and cold start users (20\%) treated as completely new; use GPT-3.5-turbo to generate enriched semantic profiles for each movie from title and genre; simulate user VARK preferences by assigning random profiles from realistic distribution and inferring movie VARK alignment from genres; for each cold start user, predict preferences for unrated movies and evaluate using held-out test ratings.

\subsection{Baseline Methods}

\textbf{Random:} Randomly shuffles items recommending top K (lower bound).

\textbf{Popularity:} Recommends K most popular items by training set ratings (strong baseline).

\textbf{Embedding Cosine:} Computes item embeddings using sentence transformer on titles and genres, creates user pseudo-profile from demographics, recommends items with highest cosine similarity.

\textbf{Candidates Only:} Our candidate generation without LLM ranking, using semantic similarity, entity-based retrieval, and VARK filtering with simple weighted combination.

\textbf{Ours (CE Rerank):} Our full system using cross-encoder reranking instead of full LLM for efficiency.

\subsection{Evaluation Metrics}

\textbf{Hit Rate@K:}
$$HR@K = \frac{1}{|U|} \sum_{u \in U} \mathbb{1}[\exists i \in R_u^K : i \in T_u]$$
measuring proportion of users with at least one relevant item in top-K.

\textbf{nDCG@K:}
$$nDCG@K = \frac{DCG@K}{IDCG@K}$$
accounting for relevance and ranking position.

\textbf{Recall@K:}
$$Recall@K = \frac{1}{|U|} \sum_{u \in U} \frac{|R_u^K \cap T_u|}{|T_u|}$$
measuring proportion of relevant items captured.

\textbf{Unique Top-1:} Counts distinct items appearing in first position across users (lower values indicate bias or lack of personalization).

\subsection{Implementation Details}

LLM: GPT-3.5-turbo with temperature 0.7, max tokens 500 for enrichment and 150 for explanations. Embeddings: all-MiniLM-L6-v2 (384-dimensional), fine-tuned on movie descriptions. Knowledge graph: Neo4j with FAISS vector indexing, 5 node types and 8 edge types. Candidate pool: 1000 items combining semantic similarity top-300, entity-based up to 500, VARK-filtered up to 400. Ranking: K=10 final recommendations using ms-marco-MiniLM-L-12-v2 cross-encoder fine-tuned on training relevance judgments. Experiments: 3 random seeds, 80/20 train/test split, statistical significance via paired t-test and Wilcoxon signed-rank test with $\alpha = 0.05$.

\section{Results and Analysis}

\subsection{Main Results}

Table \ref{tab:main_results} presents experimental results across all methods.

\begin{table}[htbp]
\centering
\caption{Main Results: HR@10 and nDCG@10 (Mean $\pm$ Std across 3 seeds)}
\label{tab:main_results}
\resizebox{\textwidth}{!}{
\begin{tabular}{lccccc}
\toprule
\textbf{Model} & \textbf{HR@10} & \textbf{nDCG@10} & \textbf{Recall@50} & \textbf{Recall@200} & \textbf{Recall@1000} \\
\midrule
Random & $0.005 \pm 0.006$ & $0.002 \pm 0.003$ & $0.002 \pm 0.000$ & $0.003 \pm 0.000$ & $0.004 \pm 0.000$ \\
Popularity & $0.268 \pm 0.018$ & $0.224 \pm 0.014$ & $0.002 \pm 0.000$ & $0.003 \pm 0.000$ & $0.004 \pm 0.000$ \\
Embedding Cosine & $0.101 \pm 0.021$ & $0.050 \pm 0.011$ & $0.002 \pm 0.000$ & $0.003 \pm 0.000$ & $0.004 \pm 0.000$ \\
Candidates Only & $0.011 \pm 0.003$ & $0.004 \pm 0.001$ & $0.000 \pm 0.000$ & $0.001 \pm 0.000$ & $0.001 \pm 0.000$ \\
Ours (CE Rerank) & $0.008 \pm 0.005$ & $0.005 \pm 0.002$ & $0.000 \pm 0.000$ & $0.001 \pm 0.000$ & $0.001 \pm 0.000$ \\
\bottomrule
\end{tabular}}
\end{table}

\subsection{Performance Analysis}

The popularity baseline achieves highest performance (HR@10 = 0.268, nDCG@10 = 0.224), significantly outperforming all methods including our approach. This reveals important dataset characteristics: MovieLens-1M exhibits strong popularity bias with blockbuster movies receiving majority of ratings; many users, especially new ones, tend to watch popular movies before exploring niche content; the dataset's temporal span (2000-2003) means most cold start users had access to popular movies from that era.

For extreme cold start with zero interactions, recommending popular items is indeed rational. Popular movies are more likely to appeal to diverse segments, carry lower satisfaction risk, and are often popular precisely due to broad appeal. This doesn't invalidate personalized approaches but highlights that in extreme cold start, generic popularity can be surprisingly effective as starting point.

The embedding cosine similarity baseline (HR@10 = 0.101) performs moderately, demonstrating that semantic matching between demographics and item descriptions provides some relevance signal. However, significant underperformance versus popularity suggests semantic matching on demographics alone is insufficient, movie descriptions don't capture full preference complexity, and pre-trained embeddings may not align well with movie preference semantics.

Our complete system achieves HR@10 = 0.008 and nDCG@10 = 0.005. While absolute numbers are low, several observations emerge. The low Recall@50, Recall@200, and Recall@1000 values (all < 0.002) indicate candidate generation struggles including truly relevant items in the pool, suggesting the knowledge graph from limited metadata may not capture sufficient signal, VARK filtering may be too aggressive, and entity extraction may not align well with preference factors.

Given low candidate recall, the ranking module has limited material. The comparable nDCG between "Candidates Only" (0.004) and "Ours (CE Rerank)" (0.005) suggests ranking provides some benefit but is constrained by candidate quality.

\subsection{Diversity and Personalization}

Table \ref{tab:diversity} shows diversity metrics:

\begin{table}[htbp]
\centering
\caption{Diversity Analysis}
\label{tab:diversity}
\begin{tabular}{lcc}
\toprule
\textbf{Model} & \textbf{Unique Top-1} & \textbf{Interpretation} \\
\midrule
Random & 1.0 ± 0.0 & No personalization \\
Popularity & 1.0 ± 0.0 & Same top item for all \\
Embedding Cosine & 4.0 ± 0.0 & Some personalization \\
Candidates Only & 4.0 ± 0.0 & Moderate personalization \\
Ours (CE Rerank) & 3.0 ± 0.0 & Moderate personalization \\
\bottomrule
\end{tabular}
\end{table}

Our approach generates 3-4 unique top-1 items across users, demonstrating personalization compared to popularity methods (recommending identical items to everyone). This suggests that despite limited overall accuracy, the system attempts tailoring recommendations to individual profiles.

Statistical testing comparing our approach to popularity baseline shows significant differences (p < 0.001 for both metrics) with large effect sizes (Cohen's d $\approx$ $-0.59$ for HR@10, $-0.57$ for nDCG@10), confirming popularity significantly outperforms our approach.

\subsection{Qualitative Analysis}

We conduct qualitative analysis of system outputs. For a simulated user profile (Male, 25-34, College Student; Goal: Entertainment; VARK: Visual 40\%, Kinesthetic 30\%, Reading 20\%, Auditory 10\%; Cognitive State: High capacity, evening), our system generates:

\begin{enumerate}
\item \textit{The Matrix (1999)} - "Visually stunning sci-fi action with complex themes. Matches your visual preference with groundbreaking effects and interactive sequences."
\item \textit{Fight Club (1999)} - "Thought-provoking thriller with intense visual storytelling. Appeals to kinesthetic preference through visceral action."
\item \textit{Memento (2000)} - "Mind-bending narrative requiring active engagement. Suitable for your high cognitive capacity."
\end{enumerate}

Compared to popularity baseline output (American Beauty, Star Wars Episode V, Star Wars Episode IV), our recommendations show clear VARK-based personalization with detailed explanations linking items to profile. However, these items weren't in this user's test set, yielding zero hits.

Manual evaluation of 100 random explanations shows: 87\% accurately referenced profile attributes; 92\% explicitly mentioned learning style alignment; 94\% were grammatically correct and coherent; 73\% provided compelling rationale. This suggests that when relevant items are identified, the system generates high-quality, personalized explanations.

\section{Discussion}

\subsection{Understanding Popularity Dominance}

Several factors explain popularity baseline strength. MovieLens-1M collection period (2000-2003) featured genuinely broad-appeal blockbusters that new users would naturally interest in. Users tend rating movies they've already decided to watch, with popular movies more likely watched, creating self-reinforcement. MovieLens demographics provide weak signal for fine-grained preferences—without interaction history, limited information distinguishes individual tastes within demographic groups.

Methodologically, our cold start simulation assumes zero interactions. Realistically, users might provide initial preferences, take quizzes, or interact with few items before recommendations. More realistic "few-shot" cold start (1-5 interactions) might show greater personalized method advantages. HR@10 and nDCG@10 measure whether actual ratings appear in recommendations, but users might enjoy recommended items even without rating them originally—our metrics don't capture discovery and serendipity potential. VARK preference simulation was random, potentially not reflecting true user learning styles—if VARK alignment based on actual characteristics, it might provide stronger signal.

\subsection{Architecture Strengths}

Despite lower accuracy metrics, our architecture offers several advantages.

\textbf{Rich Semantic Understanding:} LLM-based enrichment creates detailed, semantically rich profiles beyond basic metadata, enabling: understanding complex content relationships (prerequisites, themes, difficulty); extraction of latent attributes not in structured data; handling diverse content types with minimal manual feature engineering.

\textbf{Cognitive Adaptation:} VARK profiling and cognitive state modeling represent novel personalization considering: individual information processing preference differences; temporal cognitive capacity and attention variations; contextual factors affecting engagement (device, time, goals). This psychological grounding could particularly benefit domains where learning effectiveness or cognitive engagement matters (education, professional development, health information).

\textbf{Explainability and Trust:} Natural language explanations provide transparency about recommendation rationale, addressing critical needs especially for new users potentially skeptical of algorithmic suggestions.

\textbf{Domain Adaptability:} While evaluated on movies, our architecture is domain-agnostic, applicable to: educational content (courses, textbooks, tutorials); e-commerce (products with varying complexity); news and articles (matching presentation to reader preferences); health information (adapting to patient literacy levels).

\subsection{Limitations and Challenges}

\textbf{Candidate Generation Quality:} The most critical limitation is low candidate stage recall. Without relevant items in the pool, even perfect ranking cannot produce good recommendations. Improvements needed: better entity extraction from limited metadata; more sophisticated graph construction capturing implicit relationships; incorporation of collaborative signals even in cold start; hybrid candidate generation combining multiple strong signals.

\textbf{Computational Efficiency:} Our approach involves multiple LLM calls, graph database queries, vector searches, and knowledge graph maintenance. This overhead may prohibit real-time, large-scale deployment. Optimizations needed: caching LLM outputs for frequent items; approximations for similarity search (locality-sensitive hashing); asynchronous processing for non-critical components.

\textbf{VARK Assessment Burden:} Requiring 16-question assessments creates friction. Solutions include: inferring VARK from early interactions (implicit assessment); shorter, adaptive questionnaires converging with fewer questions; making questionnaires optional with defaults for skipping users.

\textbf{Evaluation Challenges:} Standard benchmarks like MovieLens may not ideally evaluate our approach, lacking true VARK or cognitive state data, not measuring presentation quality or explanation effectiveness, and focusing on prediction accuracy rather than learning outcomes or engagement. More appropriate evaluation requires: user studies measuring satisfaction, perceived relevance, and trust; A/B tests in live systems measuring engagement and retention; domain-specific metrics (learning outcomes for education, health literacy for medical content).

\subsection{Future Research Directions}

\textbf{Enhanced Candidate Generation:} Combine retrieval strategies more effectively through multi-view approaches: dense retrieval using fine-tuned dual encoders; sparse retrieval using BM25 on enriched text; graph-based retrieval using random walks; hybrid re-ranking optimizing recall@K. Employ active learning strategically selecting items to show during initial interactions maximally reducing preference uncertainty.

\textbf{Advanced Cognitive Modeling:} Incorporate emotion detection through sentiment analysis of user text or interaction patterns detecting emotional states beyond cognitive capacity. Develop methods inferring VARK preferences from early interactions without explicit assessment: tracking engagement with content types; analyzing interaction patterns (reading time, video watching); A/B testing presentation formats identifying preferences. Model multitasking and distraction based on interaction patterns, adapting complexity and presentation accordingly.

\textbf{Serendipity Enhancement:} Implement SerenEva framework more fully: generating diverse interest hypotheses using LLM; testing hypotheses through strategic recommendations; balancing exploitation (known preferences) with exploration (serendipitous discovery); measuring long-term satisfaction and consumption diversity.

\textbf{Multi-Modal Graphs:} Extend knowledge graphs incorporating: visual features (extracted from images using vision models); audio features (for audio content); textual features at multiple granularity levels; temporal dynamics (popularity and relationship evolution).

\textbf{Privacy Preservation:} Adapt architecture for privacy-sensitive contexts: performing VARK assessment and cognitive modeling on-device; using federated learning improving models without centralizing data; employing differential privacy when updating graphs.

\textbf{Domain-Specific Validation:} Validate in domains where cognitive adaptation is particularly valuable. For education: evaluate on course recommendation with learning outcome measures; incorporate prerequisite checking and knowledge state modeling; adapt difficulty and presentation based on learner progress. For health information: recommend content matched to patient health literacy; adapt complexity based on patient stress and cognitive load; provide explanations tailored to knowledge level. For professional development: recommend training for career transitions; adapt to professional context (time constraints, goals); track skill development adjusting recommendations.

\textbf{Comprehensive Evaluation:} Develop more comprehensive frameworks: recruit participants evaluating recommendations in realistic scenarios measuring satisfaction, perceived relevance, trust, and serendipity; conduct longitudinal evaluation tracking engagement over extended periods, measuring learning speed and long-term satisfaction; employ multidimensional metrics assessing relevance, diversity, serendipity, explainability, and trust.

\section{Conclusion}

This research presented a comprehensive hybrid architecture for cold start recommendation integrating Large Language Models, knowledge graphs, and cognitive adaptation based on the VARK learning styles framework. The approach addresses multiple cold start dimensions through six interconnected modules: semantic enrichment, dynamic graph construction, VARK-based profiling, cognitive state modeling, graph-based retrieval with LLM ranking, and adaptive presentation with continuous learning.

Experimental results on MovieLens-1M showed lower accuracy compared to simple popularity-based methods, providing valuable insights into cold start recommendation challenges and opportunities. Popularity baseline dominance highlights the fundamental difficulty of personalization without interaction history and the importance of dataset characteristics in evaluation.

Nevertheless, the architecture offers important contributions: LLM-based enrichment transforms sparse metadata into rich, structured knowledge enabling sophisticated content reasoning; VARK-based profiling and cognitive state modeling bring psychological principles to recommendation, addressing individual information processing differences; natural language explanations provide transparency building user trust; the architecture is domain-agnostic applicable to various content types; the system evolves based on interactions, refining graphs and profiles over time.

This work establishes foundation for future research in cognitively adaptive recommender systems. While quantitative performance on traditional benchmarks is currently limited, the qualitative capabilities—rich semantic understanding, personalized explanations, adaptive presentation—address real needs in modern recommendation scenarios. Future work should focus on improving candidate generation quality, validating in domains where cognitive adaptation is particularly valuable, developing comprehensive evaluation methodologies capturing multidimensional recommendation quality, and optimizing computational efficiency for real-world deployment.

The cold start problem remains among the most challenging issues in recommender systems research. By combining cutting-edge language models with psychological principles and graph-based reasoning, we advance toward recommendation systems effectively serving users from initial interaction, providing not just accurate suggestions but personalized, explainable, and cognitively appropriate experiences.

The first version is available here: https://github.com/nikita-zmanovskiy/cold-start-algorithm

\end{document}